\pdfoutput=1

\documentclass[letterpaper, 10pt, conference]{ieeeconf}  %

\IEEEoverridecommandlockouts                              %

\let\labelindent\relax                                    %

\usepackage{graphicx}            %
\usepackage{amsmath}             %
\usepackage{amssymb}             %
\usepackage{amsfonts}            %
\usepackage{enumitem}            %
\usepackage{multirow}            %
\usepackage{siunitx}             %
\usepackage{url}                 %
\usepackage{xspace}              %
\usepackage[T1]{fontenc}         %
\usepackage[hidelinks]{hyperref} %

\graphicspath{{images/}}
\DeclareGraphicsExtensions{.pdf,.png,.jpg,.jpeg}

\newcommand{\degreem}{^{\circ}} %

\newcommand{\seclabel}[1]{\label{sec:#1}}

\newcommand{\figlabel}[1]{\label{fig:#1}}
\newcommand{\tablabel}[1]{\label{tab:#1}}

\newcommand{\secref}[1]{Section~\ref{sec:#1}\xspace}

\newcommand{\figref}[1]{Fig.~\ref{fig:#1}\xspace}
\newcommand{\tabref}[1]{Table~\ref{tab:#1}\xspace}

\newcommand{\nop}{NimbRo\protect\nobreakdash-OP\xspace}
\newcommand{\dop}{DARwIn\protect\nobreakdash-OP\xspace}
\newcommand{\rop}{ROBOTIS OP2\xspace}
\newcommand{\nao}{Nao\xspace}
\newcommand{\cm}{CM730\xspace}
\newcommand{\itwoc}{I\textsuperscript{2}C\xspace}
\newcommand{\igus}{igus\textsuperscript{\tiny\circledR}\xspace}
\newcommand{\iguhop}{igus\textsuperscript{\tiny\circledR}$\!$ Humanoid Open Platform\xspace}
\newcommand{\iguhopp}{igus\textsuperscript{\tiny\circledR} Humanoid Open Platform\xspace}
\newcommand{\cpp}{C\texttt{\nolinebreak\hspace{-.05em}+\nolinebreak\hspace{-.05em}+}\xspace}

\newcommand{\term}[1]{\emph{#1}\xspace}
\newcommand{\degree}{$\degreem$\xspace}

\title{\LARGE \bf Child-sized 3D Printed igus Humanoid Open Platform}

\author{Philipp Allgeuer, Hafez Farazi, Michael Schreiber and Sven Behnke%
\thanks{All authors are with the Autonomous Intelligent Systems (AIS) Group, Computer Science Institute VI,
        University of Bonn, Germany. Email: {\tt\small pallgeuer@ais.uni-bonn.de}. This work was partially
        funded by grant BE 2556/10 of the German Research Foundation (DFG).}}

\usepackage{eso-pic}

\AtBeginDocument{\AddToShipoutPictureFG*{\AtTextUpperLeft{\put(0,\LenToUnit{9pt}){\parbox{\textwidth}{\centering\bfseries
International Conference on Humanoid Robots (Humanoids), Seoul, Korea, 2015
}}}}}
\begin{document}

\bstctlcite{IEEEexample:BSTcontrol}

\maketitle
\thispagestyle{empty}
\pagestyle{empty}

\begin{abstract}
The use of standard platforms in the field of humanoid robotics can accelerate 
research, and lower the entry barrier for new research groups. While many 
affordable humanoid standard platforms exist in the lower size ranges of up to 
60\,cm, beyond this the few available standard platforms quickly become 
significantly more expensive, and difficult to operate and maintain. In this 
paper, the \iguhopp is presented---a new, affordable, versatile and easily 
customisable standard platform for humanoid robots in the child-sized range. At 
90\,cm, the robot is large enough to interact with a human-scale environment in 
a meaningful way, and is equipped with enough torque and computing power to 
foster research in many possible directions. The structure of the robot is 
entirely 3D printed, allowing for a lightweight and appealing design. The 
electrical and mechanical designs of the robot are presented, and the main 
features of the corresponding open-source ROS software are discussed. The 3D CAD 
files for all of the robot parts have been released open-source in conjunction 
with this paper.
\end{abstract}

\section{Introduction}
\seclabel{introduction}

The field of humanoid robotics is enjoying increasing popularity, with many 
research groups from around the world having developed platforms of all sizes 
and levels of complexity to investigate challenges such as bipedal walking, 
environmental perception, object manipulation, and human-machine interaction. 
The nature of robots with humanlike kinematics gives them the mobility and 
versatility to act in everyday environments, including eventually in the home, 
in a similar way to their human counterparts. The effort to start humanoid 
robotics research on a real platform can be high however, and often large 
amounts of time need to be spent on design, construction, firmware and software 
before the platform is sufficiently functional that the real research 
topics can be addressed. This is a barrier of time, money, and multifaceted 
technical expertise that can inhibit many research groups from gaining entry 
into the humanoid field. The availability of standard robot platforms can 
significantly reduce these initial hurdles, facilitating wider dissemination of 
humanoid robots, and greater collaboration and code exchange between groups that 
share a common platform. Capable standard platforms have the ability to 
invigorate and accelerate academic research, as researchers can then focus 
chiefly on the challenges that are most interesting to them, and contribute 
their advances to the state of the art in their field of specialisation.

\begin{figure}[!t]
\parbox{\linewidth}{\centering
\raisebox{8pt}{\includegraphics[width=0.48\linewidth]{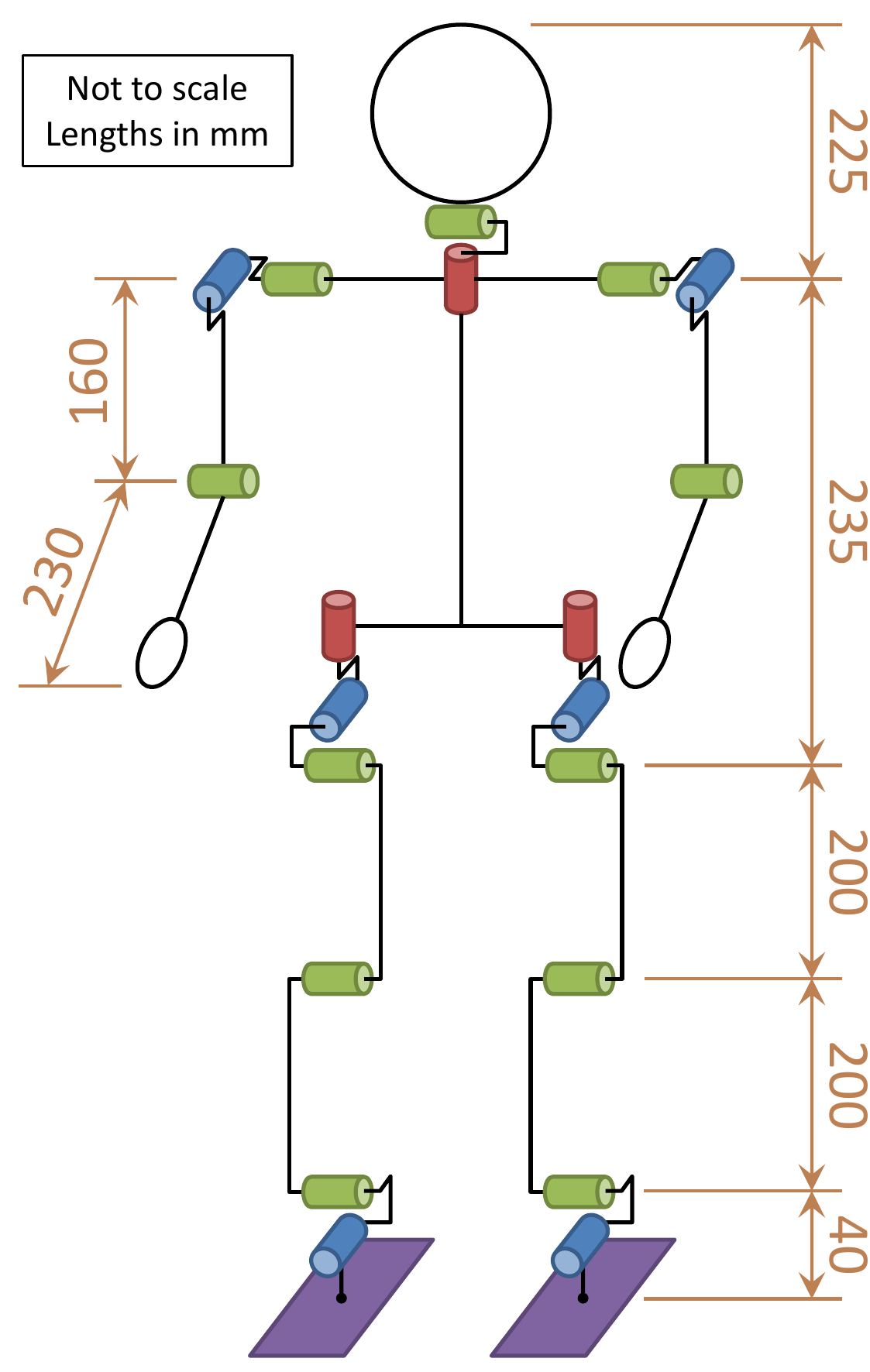}}\hspace{0.04\linewidth}\includegraphics[width=0.48\linewidth]{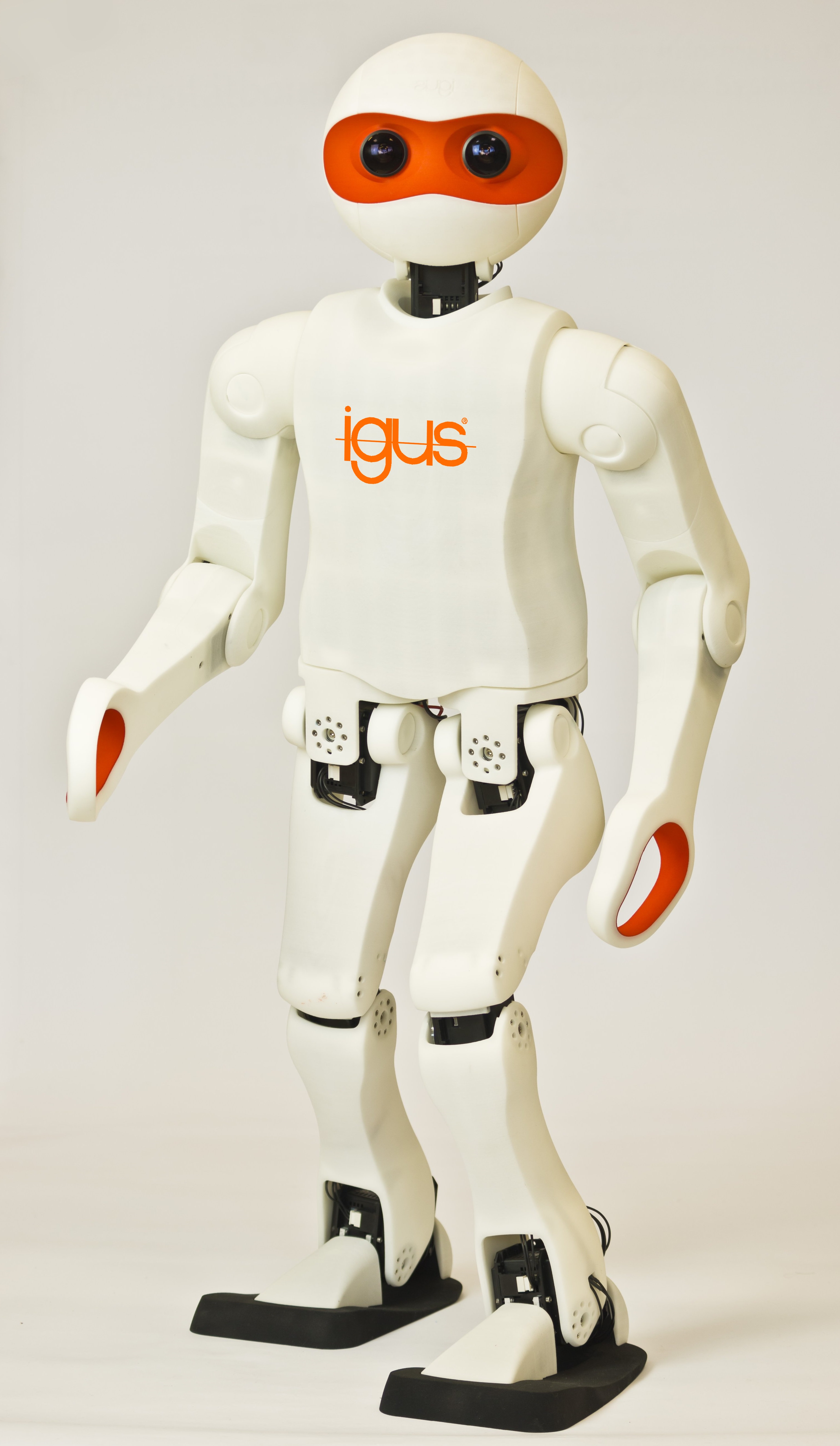}}
\caption{The \iguhop and its kinematics.}
\figlabel{P1_teaser}
\end{figure}

A number of standard humanoid robot platforms have been developed over the 
years, including notably the Aldebaran \nao and the Robotis \dop, both of which 
have seen much success. The \iguhop, introduced in this paper, is shown in 
\figref{P1_teaser}. The platform is a collaboration between researchers at the 
University of Bonn and \igus GmbH, a leading manufacturer of polymer bearings 
and energy chains. The robot has already been demonstrated at numerous 
industrial trade fairs, in addition to demonstrations at the most recent RoboCup 
and RoboCup German Open. The \iguhop seeks to close the gap between small, 
albeit affordable, standard humanoid platforms, and larger significantly more 
expensive ones such as the Honda Asimo and Boston Dynamics Atlas robots. By 
developing this platform, we seek to enable teams to work with an affordable 
robot of a size large enough to interact meaningfully with the environment. 
Furthermore, we designed the platform to be as open, modular, maintainable and 
customisable as possible, to allow the robot to be adapted to a variety of 
research tasks with minimal effort. The choice of using almost exclusively 3D 
printed plastic parts for the mechanical components of the robot forms the core 
of this philosophy, and also greatly simplifies the manufacture of the robots. 
As a result, individual parts can easily be modified, reprinted and replaced to 
augment the capabilities of the robot. For example, if a gripper was to be 
required for an application, it would be easy to design a replacement lower arm 
part to accommodate this. As a further example of the extensibility of the 
platform, the internal PC also provides the appropriate interfaces to 
incorporate a microphone and speakers, if this is required. In consonance with 
our aims of producing a platform that is entirely open, the ROS middleware 
\cite{Quigley2009} was chosen as the basis of the software developed for the 
\iguhopp. This promotes the modularity, visibility, reusability, and to some 
degree also the platform independence, of the produced robot software in this 
research community-driven ecosystem. The complete hardware and software designs, 
the former in the form of print-ready 3D CAD files \cite{IguhopHardware} and the 
latter with documentation \cite{IguhopSoftware}, are available open source.

\section{Related Work}
\seclabel{related_work}

\begin{figure}[!t]
\parbox{\linewidth}{\centering\includegraphics[height=8.0cm]{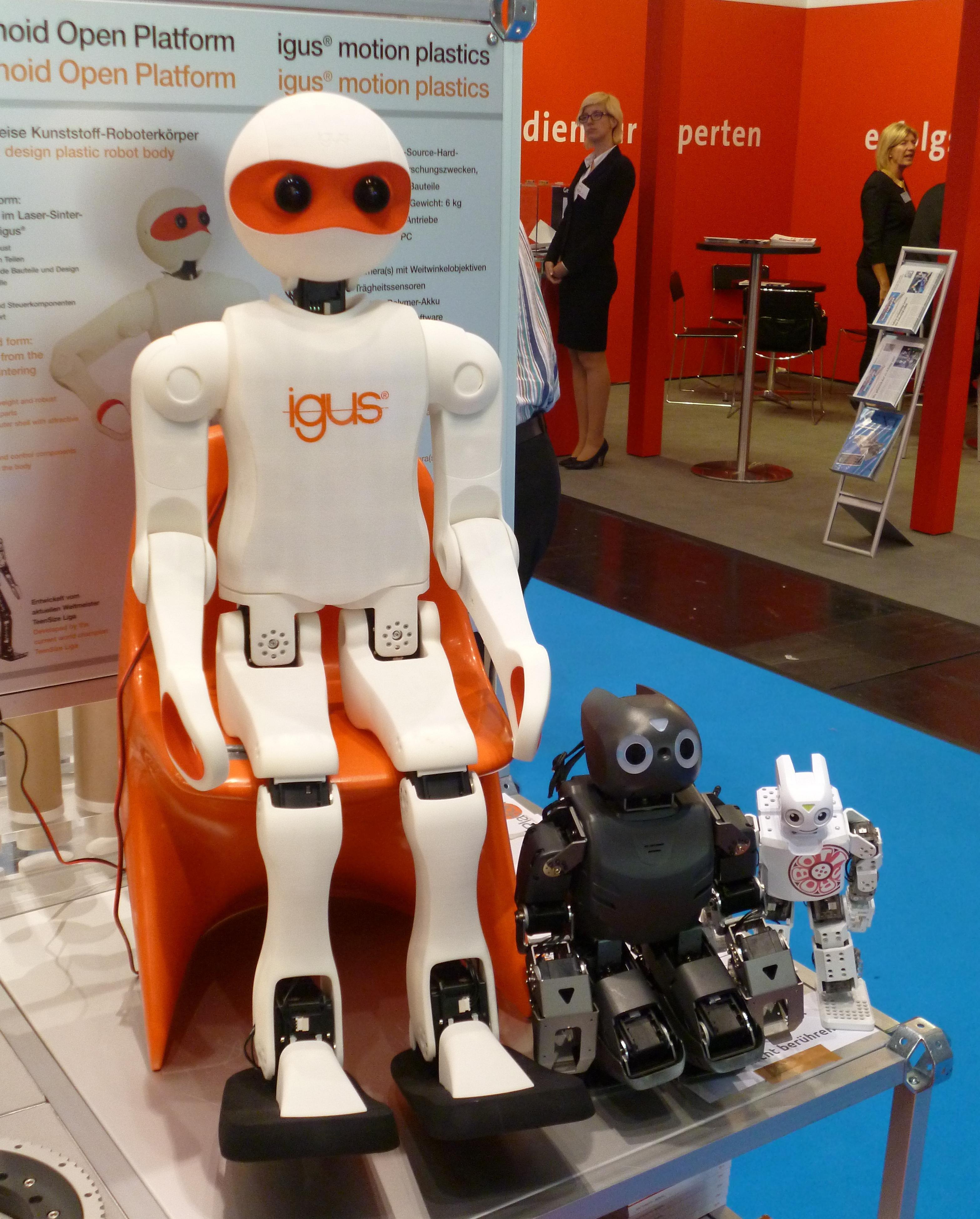}}
\caption{The \iguhop in size comparison with the Robotis \dop and Darwin-Mini robots.}
\figlabel{P1_comparison}
\end{figure}

Over the last decade, a number of humanoid robotic platforms have been 
introduced that demonstrate how the use of a standard platform can accelerate 
research and development, by alleviating the need for teams to each `reinvent 
the wheel' for low level tasks such as communications and walking. The most 
prominent example of this is the \nao robot \cite{Gouaillier2009}, developed by 
Aldebaran Robotics and first released publicly in 2008. By now, many 
thousands of \nao robots are in use all over the world. The large dissemination 
was driven in part by their use as the standard humanoid platform for the 
RoboCup Soccer SPL competition. The \nao comes with a rich set of features, such 
as a variety of available gaits, a programming SDK, and well-developed 
human-machine interaction components. However, at \SI{58}{cm} tall, \nao{}s 
often require special miniaturised environments, with objects on a scale that 
the robot can manipulate or step onto. Also, as a proprietary product, there are 
only very limited possibilities for own hardware repair and customisation.

Another example of a successful standard platform is the \dop \cite{Ha2011}, and 
its recent successor the \rop, distributed by Robotis. Both robots are quite 
similar in design and architecture, and stand at \SI{45.5}{cm} tall, almost 
exactly half the size of the \iguhop. \figref{P1_comparison} shows a size 
comparison between the two robots and a Darwin-Mini, which is only \SI{27}{cm} 
tall. In contrast to the closed proprietary design of the \nao robot, the \dop 
was designed as an open platform that allows users to operate, maintain and 
customise the robot as they desire. Nevertheless, the size of the robot has 
remained a limiting factor to its range of applications.

Two more recently developed platforms include the Intel Jimmy robot, and the 
Poppy robot from the Inria Flowers Laboratory \cite{Lapeyre2014}. Both of these 
robots are open source platforms that are for the most part 3D printed. At 
\SI{65}{cm} tall, the Jimmy robot is intended for social interactions, and comes 
with software based on the \dop framework that includes for example a walking 
engine. The robot only uses 3D printed parts for its outer shell, but its weight 
is supported by an internal aluminium frame. The Poppy robot on the other hand, 
is more completely 3D printed, and is intended as a research platform with a 
compliant bio-inspired morphology. The robot is \SI{84}{cm} tall, has a 
multi-articulated trunk, and features a skeletonised design. Poppy has so far 
only demonstrated walking in an assisted manner, and is generally intended for 
non-autonomous use, with persistent cabled connections to the robot and 
off-board processing.

Larger standard platforms, such as the Asimo \cite{Hirai1998}, HRP 
\cite{Kaneko2009} and Atlas robots, quickly run into the limitation that they 
become an order of magnitude more expensive, and more difficult, or even 
dangerous, to operate and maintain. Such large robots also have a significantly 
lower robustness to falling in terms of hardware damage, and require a gantry in 
normal use. These factors limit the possibility of the use of such robots by 
most research groups.

The \iguhop bridges the gap between the existing larger and smaller standard 
platforms in that it is large enough to act in a human-scale environment, yet 
small enough to fall down and get back up, and lighter for its size than all of 
the other robots when compared by body mass index (BMI). The cost of the 
platform is about twice that of the \dop. In 2012, work 
commenced on the first prototype of a child-sized humanoid platform, dubbed the 
\nop \cite{Schwarz2012}. The \nop was constructed mainly out of aluminium 
profiles and carbon composite sheets, with only the head and a few other small 
parts being 3D printed. The manufacturing was completed in-house, and required 
the milling of parts from up to four sides, but yielded a very rigid and 
light-weight result. An earlier form of the ROS software was developed for the 
\nop, and both the hardware and software were released open source 
\cite{Allgeuer2013a}. This produced quite some interest in the platform, and 
although it should be stated that it was an academic as opposed to a commercial 
effort, several groups from around the world purchased the platform, or produced 
variants of their own using our hardware specifications. The \iguhop represents 
the next step in the evolution of the \nop into a robust multifaceted humanoid 
robotics platform, and is to be seen as an open contribution to the humanoid 
robotics community to help more research groups enter the field.

\section{Robot Design Concept}
\seclabel{design_overview}

\begin{table}
\renewcommand{\arraystretch}{1.3}
\caption{\vspace{0.7ex}\iguhop specifications\vspace{-0.7ex}}
\tablabel{P1_specs}
\centering
\begin{tabular}{c c c}
\hline
\textbf{Type} & \textbf{Specification} & \textbf{Value}\\
\hline
\multirow{5}{*}{\textbf{General}} & Height & \SI{90}{cm}\\
& Weight & \SI{6.6}{kg}\\
& Battery & 4-cell LiPo (\SI{14.8}{V}, \SI{3.8}{Ah})\\
& Battery Life & \SI{15}{}--\SI{30}{\minute}\\
& Material & Polyamide 12 (PA12)\\
\hline
\multirow{7}{*}{\textbf{PC}} & Product & Gigabyte Brix GB-BXi7-5500\\
& CPU & Intel i7-5500U (4 threads)\\
& Frequency & \SI{2.4}{}--\SI{3.0}{GHz}\\
& RAM & \SI{4}{GB} DDR3\\
& Disk & \SI{120}{GB} SSD\\
& Network & Ethernet, Wi-Fi, Bluetooth\\
& Other & 4$\,\times\,$USB 3.0, HDMI, MiniDP\\
\hline
\multirow{4}{*}{\textbf{\cm}} & Microcontroller & STM32F103RE (Cortex M3)\\
& Memory & \SI{512}{KB} Flash, \SI{64}{KB} SRAM\\
& Frequency & \SI{72}{MHz}\\
& Other & 3$\,\times\,$Buttons, 7$\,\times\,$Status LEDs\\
\hline
\multirow{4}{*}{\textbf{Actuators}} & Total & 8$\,\times\,$MX-64, 12$\,\times\,$MX-106\\
& Head & 2$\,\times\,$MX-64\\
& Each Arm & 3$\,\times\,$MX-64\\
& Each Leg & 6$\,\times\,$MX-106\\
\hline
\multirow{6}{*}{\textbf{Sensors}} & Encoders & \SI{4096}{ticks/rev}\\
& Gyroscope & 3-axis (L3G4200D chip)\\
& Accelerometer & 3-axis (LIS331DLH chip)\\
& Magnetometer & 3-axis (HMC5883L chip)\\
& Camera & Logitech C905 (720p)\\
& Camera Lens & Wide-angle lens with 150\degree\!FOV\\
\hline
\end{tabular}
\end{table}

A summary of the main hardware specifications of the \iguhop is shown in 
\tabref{P1_specs}. Other than an appealing overall aesthetic appearance, for which a 
design bureau was engaged, the main criteria for the design were the simplicity 
of manufacture, assembly, maintenance and customisation. To satisfy these 
criteria, a modular design approach was used. This is evident, for example, in 
the choice of using a high performance small form factor PC with standard 
mounting points in a location in the torso where space requirements are 
flexible. As such, it is very simple to upgrade the PC in the robot. If the 
layout of the external PC sockets changes, the only potentially required 
hardware change is the replacement of the white plastic cover on the back of the 
robot that covers the PC. As this cover is printed in a modular fashion 
separately from the remainder of the torso, it is very easily replaced 
with a part that has holes in the correct locations for the new PC. Since the 
initial \nop prototype was built, the PC has been changed twice to augment the 
computational power.

The modular design of the robot is also demonstrated by its ability to allow 
quick design iterations. Due to the 3D printed nature of the robot, parts can be 
modified and replaced with great freedom. For instance, in one of the first 
iterations of the design, it was found that one of the leg parts did not allow 
for an adequate range of motion in the hips. This issue was quickly resolved 
with the simple change of the associated dimensions in the CAD model, and the 
subsequent reprinting of the part.

\section{Mechanical Design}
\seclabel{mechanical}

While the kinematic structure of the \iguhop is very similar to that of the 
\nop, its structural design was changed fundamentally. This was 
motivated largely by the desire for a greater overall visual and aesthetic 
appeal of the robot, but it also had many other benefits, such as a significant 
reduction in the total number of parts. The resulting robot design, shown in 
\figref{P1_teaser}, was awarded the first RoboCup Design Award in 2015, based on 
criteria such as performance, simplicity and ease of use. Whereas the \nop 
consisted entirely of a strongly connected array of load-bearing aluminium and 
carbon composite structures with no outer facade or walls, the new design 
consists entirely of the uniform white plastic exoskeleton that is visible 
externally. This facilitated a transition to cable routing internally through 
the limbs, as opposed to the partially external cable routing that was utilised 
in the \nop. The plastic parts of the new design are 3D printed from Polyamide 
12 in increments of less than \SI{0.1}{mm} using a Selective Laser Sintering 
(SLS) process. It is important to note that there are no extra parts behind the 
outer surface that help support the robot structure. The exoskeleton is 
simultaneously load-bearing and for outward appearance, and strongly fulfils 
both these functions. This allows for dramatic space and weight savings, as 
evidenced by the \iguhop's very low weight. The structural integrity and 
resistance to deformation and buckling is ensured through modulation of the wall 
thickness in the areas that require it, and through strategic widespread use of 
ribs and other geometric strengthening features, which are printed directly as 
part of the exoskeleton. Essentially, through the freedoms of 3D printing, the 
plastic part strengths can be assigned exactly where they are needed, and not 
unnecessarily in other locations. This is a second reason why the weight of the 
new mechanical design is so low. If a weak spot is identified through practical 
experience, as indeed happened during walking tests, the parts can locally be 
strengthened in the CAD design without significantly impacting the remainder of 
the design.

The kinematic structure of the robot is shown in \figref{P1_teaser}. Starting 
from the trunk, the head can first yaw and then pitch, and the arms first pitch 
and then roll in the shoulder, before allowing pitch again in the elbow joint. 
There are three degrees of freedom (DOF) in each hip---first yaw, then roll and 
then pitch---one DOF in each knee, and two DOF in each ankle---first roll and 
then pitch. Position controlled Robotis Dynamixel MX-64 and MX-106 servos are 
used for all of the actuators, and mechanically form the sole connections between 
the parts, except for in the hips, where \igus axial thrust bearings provide 
the required dry rubbing self-lubrication.

Two noteworthy changes in the kinematics over the \nop, are that the 
feet are more strongly reinforced---increasing their rigidity to elastic 
deformation---and that the mount point for the ankle is further back on the 
foot. This reduces the effect that the backlash in the ankle pitch actuator has 
on the stability and balance of the robot during walking, and avoids similar 
effects caused by bending of the feet under unbalanced loads.

\section{Electronics}
\seclabel{electrical}

Other than for an upgrade of the PC, the electrical design of the \iguhop has 
remained very similar to that of the \nop prototype, which in turn was inspired 
by the configuration of the \dop robot. A block diagram of the electrical 
subsystem of the robot design is shown in \figref{electrical_architecture}. At 
the heart of the electrical design is the Robotis \cm sub controller board, 
which contains power management features, management of a Dynamixel bus, and a 
Cortex M3 STM32F103RE microcontroller running at \SI{72}{MHz}, alongside 
numerous other peripheral features. The \cm is connected via USB to a PC running 
a full 64-bit Ubuntu operating system. All of the robot control software and 
processing runs on this Intel i7-5500U PC. DC power is provided to the system 
via a power board that incorporates a switch for the entire robot. One or both 
DC power and a 4-cell Lithium Polymer (LiPo) battery can be connected, and the 
higher voltage of the two is forwarded to supply all electronics of the robot.

The main purpose of the \cm is to electrically interface the eight MX-64 and 
twelve MX-106 servos, all connected on a single Dynamixel bus. In the \iguhop 
design, the use of servo daisy-chaining has been kept to a minimum for stability 
and robustness reasons. Five separate connectors on the side of the \cm are used 
to create a star topology, with daisy-chaining only being required for 
mechanical reasons in the elbow and ankle joints. The actuators are used in 
position control mode, and provide joint encoder feedback with a resolution of 
4096 ticks per revolution via the Dynamixel bus.

\begin{figure}[!t]
\parbox{\linewidth}{\centering\includegraphics[width=1.0\linewidth]{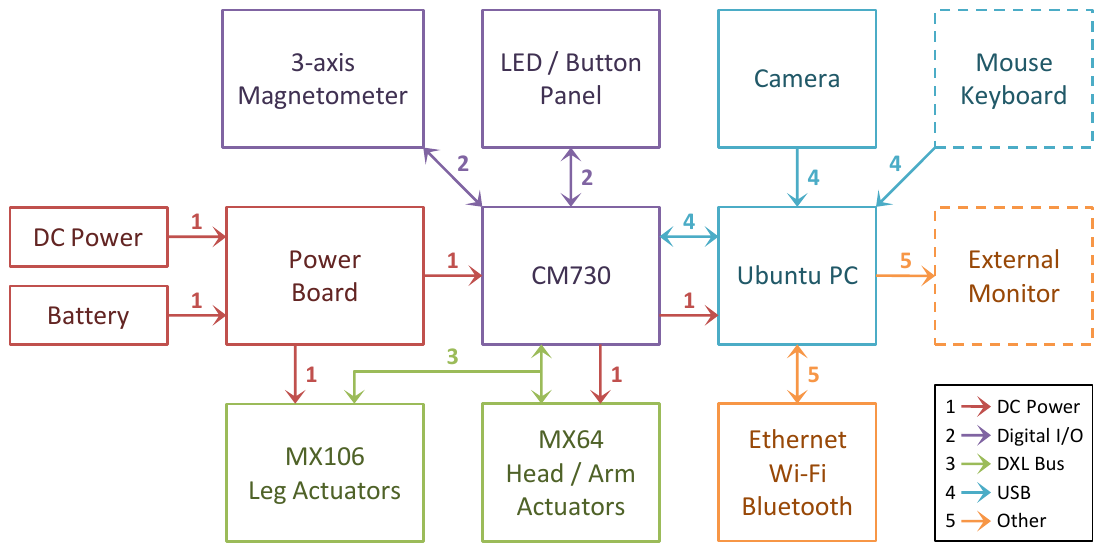}}
\caption{System block diagram of the electronic components and connections in the \iguhop.}
\figlabel{electrical_architecture}
\end{figure}

Due to a number of factors, including reliability, performance, data latency, 
throughput and changed electrical connections, the standard shipped firmware of 
the \cm did not satisfy the requirements of the \iguhop. As a result, the 
firmware of the \cm was fundamentally redesigned and completely rewritten. Many 
changes and improvements were made in the process, the most notable of these 
being an extension of the Dynamixel protocol for communications between the \cm 
and the PC. These extensions translated into significant gains in bus 
stability and error tolerance, as well as in time savings for bulk reads of 
servo data. It should be noted however, that the upgraded communications 
protocol is fully back-compatible with the standard Dynamixel protocol.

In addition to the actuators, the \cm also connects to an interface panel that 
contains three buttons, five LEDs and two RGB LEDs. These elements are managed 
by the PC, and can be used to display information about the 
internal state of the robot, and provide control triggers. The \cm incorporates 
a 3-axis gyroscope and a 3-axis accelerometer. An additional 3-axis magnetometer 
is connected via an \itwoc interface to the onboard microcontroller. The 
resulting 9-axis IMU is polled at high frequency and placed into registers 
that can be queried by the PC. A 720p Logitech C905 USB 
camera, located in the head of the robot and fitted with a 150\degree FOV 
wide-angle lens, is connected directly to the PC. Further available external 
connections to the PC include USB, HDMI, Mini DisplayPort, Gigabit Ethernet,
IEEE 802.11b/g/n Wi-Fi, and Bluetooth 4.0. By default, the 
Ethernet network interface of the PC is configured system-wide to automatically 
switch between static and DHCP connections as required, also allowing 
simultaneous use of the static secondary IP address while DHCP is active.

\section{Software Architecture}
\seclabel{software}

\begin{figure}[!t]
\parbox{\linewidth}{\centering\includegraphics[width=1.0\linewidth]{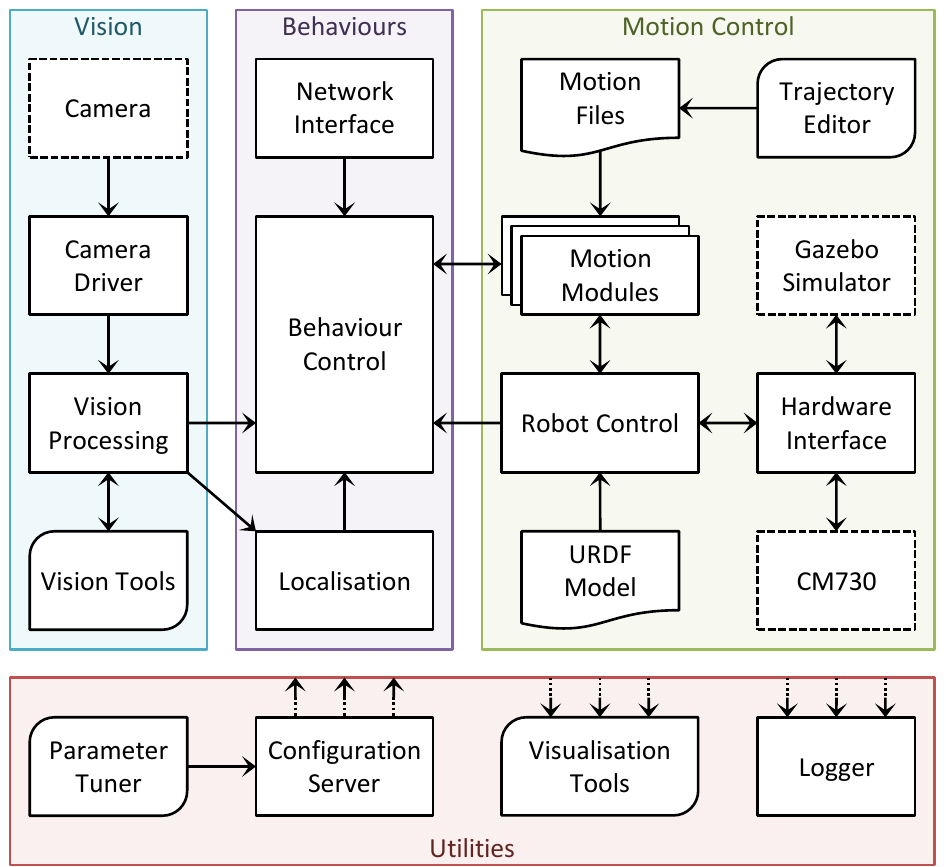}}
\caption{Architecture of the \iguhop ROS software.}
\figlabel{software_architecture}
\end{figure}

The software that has been developed for the \iguhop is a continuous evolution 
of the \cpp ROS-based software that was written for the \nop 
\cite{Allgeuer2013a}, and is available open source \cite{IguhopSoftware}. The 
overall software architecture, illustrated in \figref{software_architecture}, 
has not fundamentally changed since the release of the \nop software. Many new 
software components have been added though, and the individual components that 
already existed have undergone vast changes and improvements to enhance 
functionality, reliability and robustness. The software was developed under the 
guise and target application of humanoid robot soccer \cite{Schwarz2013}, but 
with suitable adaptation of the behaviour control, vision processing and motion 
module sections, software for virtually any other application can be realised. 
This is possible because of the strongly modular way in which the software was 
written, greatly supported by the natural modularity of ROS. In many 
parts of the software framework, like for example in the choice of gait, plugin 
schemes are used for exactly this purpose, and individual tasks have been 
separated as much as possible into different nodes, often with great 
independence.

At the heart of the lower level control section of the \iguhop is the 
\term{Robot Control} ROS node, which runs a hard real-time control loop that 
manages sensor and actuator communications, state estimation and motion 
generation. A URDF (Unified Robot Description Format) model of the robot is 
loaded from file, and a per-launch configurable selection of motion module 
plugins use this information in addition to sensory perception to generate 
dynamic motions for the robot. A further hardware interface plugin scheme is 
used to manage the physical communications with the robot, or interfaces to
a virtual dummy robot or a physically simulated one in the 
Gazebo simulator. The higher level planning and artificial intelligence is 
implemented in a \term{Behaviour Control} ROS node, based on the hierarchical 
Behaviour Control Framework \cite{Allgeuer2013}. Processing of the camera images 
is implemented in a separate vision processing ROS node, which publishes the 
information required by the behaviours and the localisation ROS nodes. A number 
of framework-wide helper nodes have also been implemented, of which the 
visualisation tools, configuration server and logger find the most prominent 
use. The configuration server is a centralised storage location and manager for 
software parameters, and replaces the ROS parameter server for dynamic 
reconfiguration of the software, and the parameter tuner GUI allows live 
modification of the parameters. Typically, the visualisation tools and parameter 
tuner are run on an external PC, but the configuration server and logger are 
run on the robot.

\begin{figure}[!t]
\parbox{\linewidth}{\centering\includegraphics[width=1.0\linewidth]{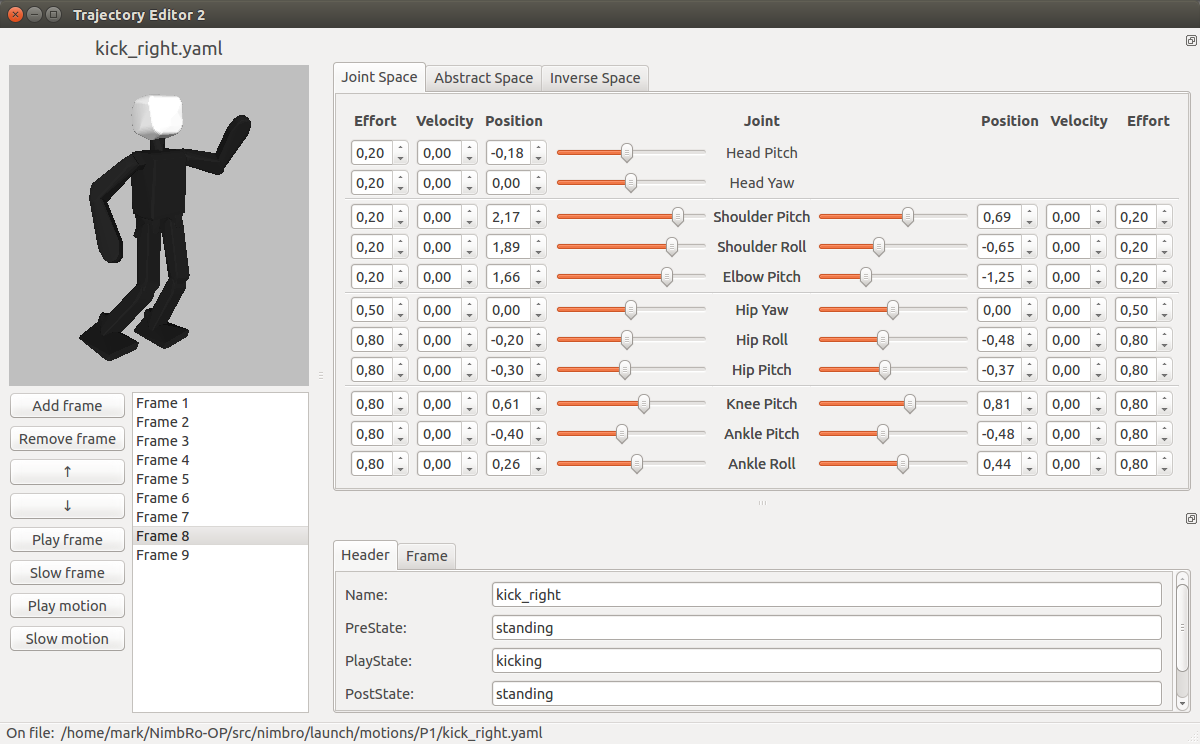}}
\caption{The trajectory editor used to design keyframe motions for the \iguhopp in the
joint, inverse and abstract control spaces.}
\figlabel{trajectory_editor}
\end{figure}
\section{Dynamic Motion Generation}
\seclabel{motion_generation}

\subsection{Compliant Actuation}
\seclabel{servo_model}

In controlling the actuators of a robot, it is desirable to have a controller 
that can follow commanded trajectories, but at the same time remain compliant 
\cite{Schwarz2013a}. In the case of the \iguhop, inverse dynamics calculations 
from the RBDL library \cite{Felis2013} are used in a feed-forward manner to 
predict the torques that are required in order to follow a particular joint 
command. These torques are then used to improve tracking performance, and allow 
more compliant settings to be used in the Dynamixel actuator control loops. 
Based on the robot state, support coefficients are estimated for each limb, and 
the inverse dynamics are used to provide gravity compensation based on these 
support coefficients via the principle of superposition. The torques required to 
counter the inertial effects derived from the required joint velocities and 
accelerations are calculated in a separate execution of the inverse dynamics. 
The total required torque is summated for each joint and sent to the servo 
control loop. The compliant servo actuation framework was found to increase 
battery life and reduce servo overheating and wear.

\subsection{State Estimation}
\seclabel{state_estimation}

Estimation of the robot state is a fundamental component of essentially all 
dynamic motion generation algorithms that implement closed loop control. 
Proprioception in the ROS software is based on the joint encoder values that are 
returned by the servos, and managed by the convenient ROS-native tf2 library, 
which uses the URDF model of the robot. Estimation of the 3D orientation of the 
robot relative to its environment is a more difficult task however, and requires 
fusion of the 9-axis IMU measurements. Based 
in part on a novel way of representing orientations, namely the \term{fused 
angles} representation \cite{Allgeuer2015}, a 3D nonlinear passive complementary 
filter was developed for this purpose \cite{Allgeuer2014}. This filter returns 
the full 3D estimated orientation of the robot, and is proven to be globally 
stable.

An immediate application of the results of the state estimation
is the fall protection motion module, which disables torque in all 
servos if the estimated angular deviation from vertical exceeds the limit of 
assumed irrecoverability. This mechanism aims to protect the servos and 
mechanical components from damage due to the impulsive impacts associated with 
falling. After falling, the state estimation is used to determine whether the 
robot needs to get up from the front, back or side.

\subsection{Keyframe Motions}
\seclabel{keyframe_motions}

There are many situations where a robot may have to execute a particular 
hand-tuned motion a number of times. In such situations, it is useful to be able 
to express and tune the motion as a set of pose keyframes that are automatically 
interpolated to become a complete whole body motion, which can be played back on 
request. This is the purpose of the motion player in the ROS software, which 
implements a generic nonlinear keyframe interpolator that, given the desired 
keyframe time intervals, is able to smoothly connect specifications of joint 
positions and velocities. In addition to this, the keyframe interpolator can 
also modulate the joint efforts and support coefficients (refer to 
\secref{servo_model}) used during the motion. This allows the compliant servo 
actuation framework to be used meaningfully during motions with changing support 
conditions.

To make the keyframe motion player practicable however, a suitable tool for the 
editing of raw motion specification files is also required. 
\figref{trajectory_editor} shows a screenshot of the trajectory editor that 
was developed for the \iguhop. With this editor, all aspects of the motion files 
can be edited in a user-friendly environment, and with an interactive 3D preview 
of the robot poses. What distinguishes this editor is that the keyframes can be 
edited either in:
\begin{itemize}
\item \term{Joint space}, by direct manipulation of the joint angles,
\item \term{Inverse space}, by specification of the limb end effector 
(i.e.~foot, hand) poses in terms of Cartesian coordinates and a 3D orientation, or
\item \term{Abstract space}, by specification of the required arm, foot and leg 
parameters.
\end{itemize}
The last of the three spaces, the \term{abstract space}, is a space that was 
developed specifically for legged robots in the context of walking and balance 
\cite{Behnke2006}. It abstracts the pose of the limb of a robot, nominally a 
leg, into specifications of the required leg extension, $x$ and $y$ rotations of 
the foot relative to the trunk, and $x$, $y$ and $z$ rotations of the leg centre 
line relative to the trunk, following the Euler ZXY convention. The \term{leg 
centre line} is defined to be the line joining the centre of the hip joint and 
the ankle joint. Basic trigonometric conversions between joint space and 
abstract space exist.

Motions that have been designed using the trajectory editor include kicking, 
waving, demonstration and get-up motions. The waving and demonstration motions 
have been used at various industrial trade fairs for publicity and interaction 
purposes. The kicking motions are relatively dynamic, and are able to propel a 
size 4 FIFA ball \SI{3.5}{\metre} across \SI{32}{mm} blade length artificial 
grass, which provides a high rolling resistance. A still image of the kicking motion 
is shown in \figref{P1_getup}. The two main get-up motions of the \iguhopp are 
also shown in the same figure. Four additional get-up motions handle the cases 
where the robot is lying on its side. With help of the support coefficients 
feature, the get-up motions can be efficient and controlled, and are 
considerably faster and more dynamic than those of most other robots of 
comparable size. In fact, there are extended leg flight phases during both the 
prone and supine get-up motions, where the only points of contact of the robot 
with the ground are the tips of the two arms, during which time the legs of the 
robot are swung underneath it. The get-up motions from the prone and supine 
positions take 3 and 4 seconds respectively, measured from the moment that the 
robot starts to fade in its torque to when the robot is supported in a balanced 
way on only its two feet.

\begin{figure}[!t]
\parbox{\linewidth}{\centering\includegraphics[width=1.0\linewidth]{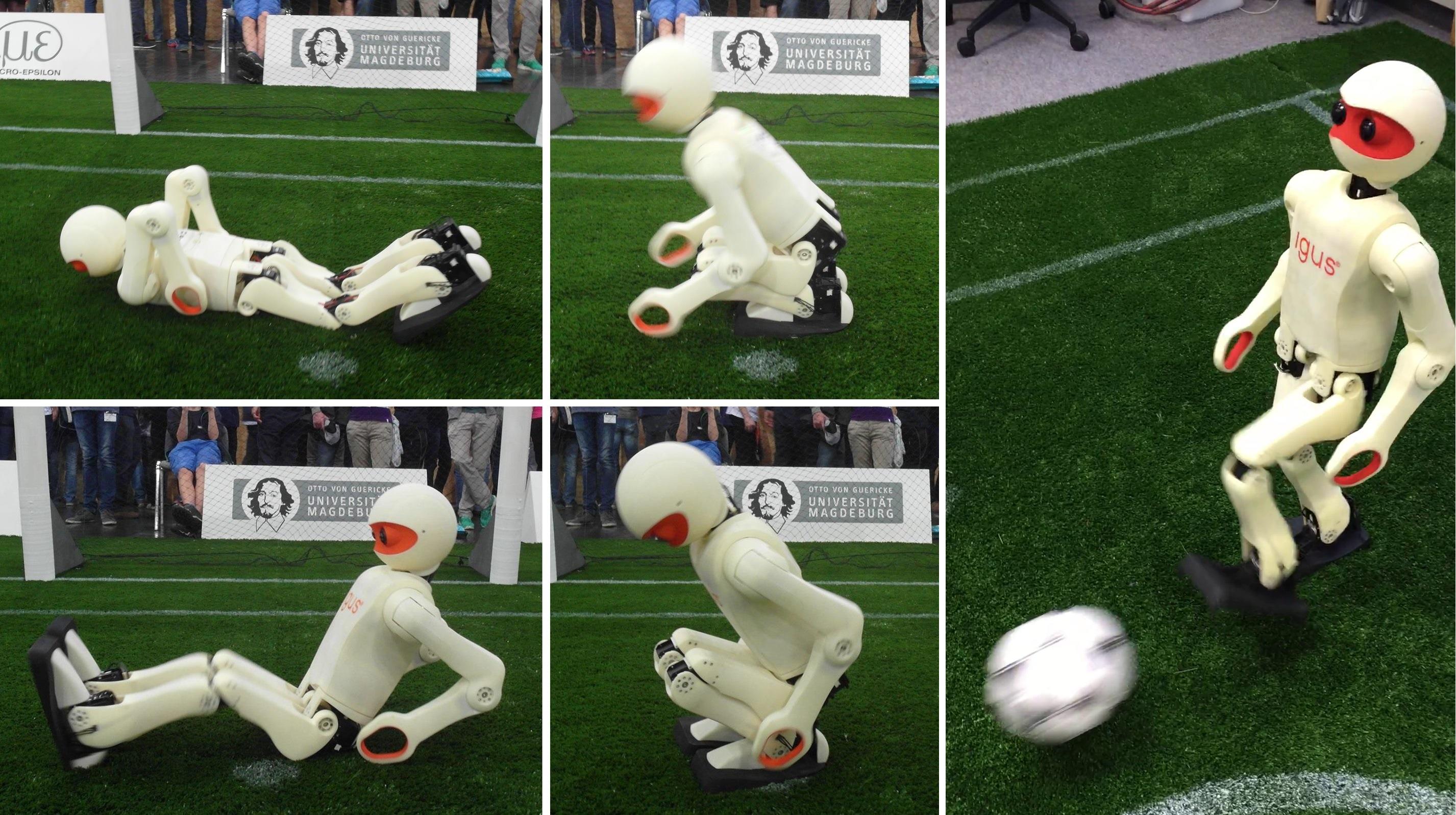}}
\caption{Dynamic get-up motions of the \iguhop, from the prone (top row) and supine
(bottom row) lying positions, and a still image of the dynamic kick motion.}
\figlabel{P1_getup}
\end{figure}
\subsection{Gait Generation}
\seclabel{gait_generation}

Motivated by the changed game environment at the RoboCup competition---the 
chosen application domain for our own use of the \iguhop---the gait generation 
has been adapted to address the new challenge of walking on artificial 
grass. The use of a soft, deformable and unpredictable walking surface imposes 
extra requirements on the walking algorithm. Removable rubber cleats have been 
added at the four corners underneath each foot of the robot to improve the grip 
on the artificial grass. This also has the effect that the ground reaction 
forces are concentrated over a smaller surface area, mitigating at least part of 
the contact variability induced by the grass.

The walking gait in the ROS software is based on an open loop central pattern 
generated core that is calculated from a gait phase angle that increments at a 
rate proportional to the desired gait frequency. This open loop gait is 
formulated in a combination of the abstract and inverse pose spaces (see 
\secref{keyframe_motions}), and extends the open loop gait of our previous work 
\cite{Missura2013a}. The main changes that have been made include:
\begin{itemize}
\item The integration of an explicit double support phase of configurable length 
for greater walking stabilisation and passive damping of oscillations,
\item The modification of the leg extension profiles to transition more smoothly 
between swing and support phases,
\item The incorporation of a sagittal leg angle term that trims the angle 
relative to the ground at which the feet are lifted during stepping,
\item The addition of a support coefficient transitioning profile for use with 
the compliant servo actuation,
\item The use of a dynamic pose blending algorithm to enable smoother 
transitions to and from walking,
\item The adaptation of the gait command velocity-based leaning strategy to use 
a hip motion instead of a leg angle motion, and
\item The introduction of a gait command acceleration-based leaning strategy.
\end{itemize}
A number of simultaneously operating basic feedback mechanisms have been built 
around the open loop gait core to stabilise the walking. The feedback in each of 
these mechanisms derives from the fused pitch and fused roll \cite{Allgeuer2015} 
state estimates.

\textbf{Virtual Slope Walking:}
If the robot is walking with a non-zero velocity in the sagittal direction and 
begins to tip either forwards or backwards, there is a possibility that due to 
the leg swing motion the swing foot unintentionally collides with the ground. 
This causes premature contact of the foot with the ground, and the robot briefly 
pushes its leg in a direction that undesirably promotes further tipping of the 
robot. Virtual slope walking adjusts the inverse kinematics height of the feet 
in proportion to their swing state and measured sagittal torso angle in such a 
way that the robot effectively lifts its feet more for foot placements if it is 
tipping in the direction in which it is walking.

\begin{figure}[!t]
\parbox{\linewidth}{\centering\includegraphics[width=0.9\linewidth]{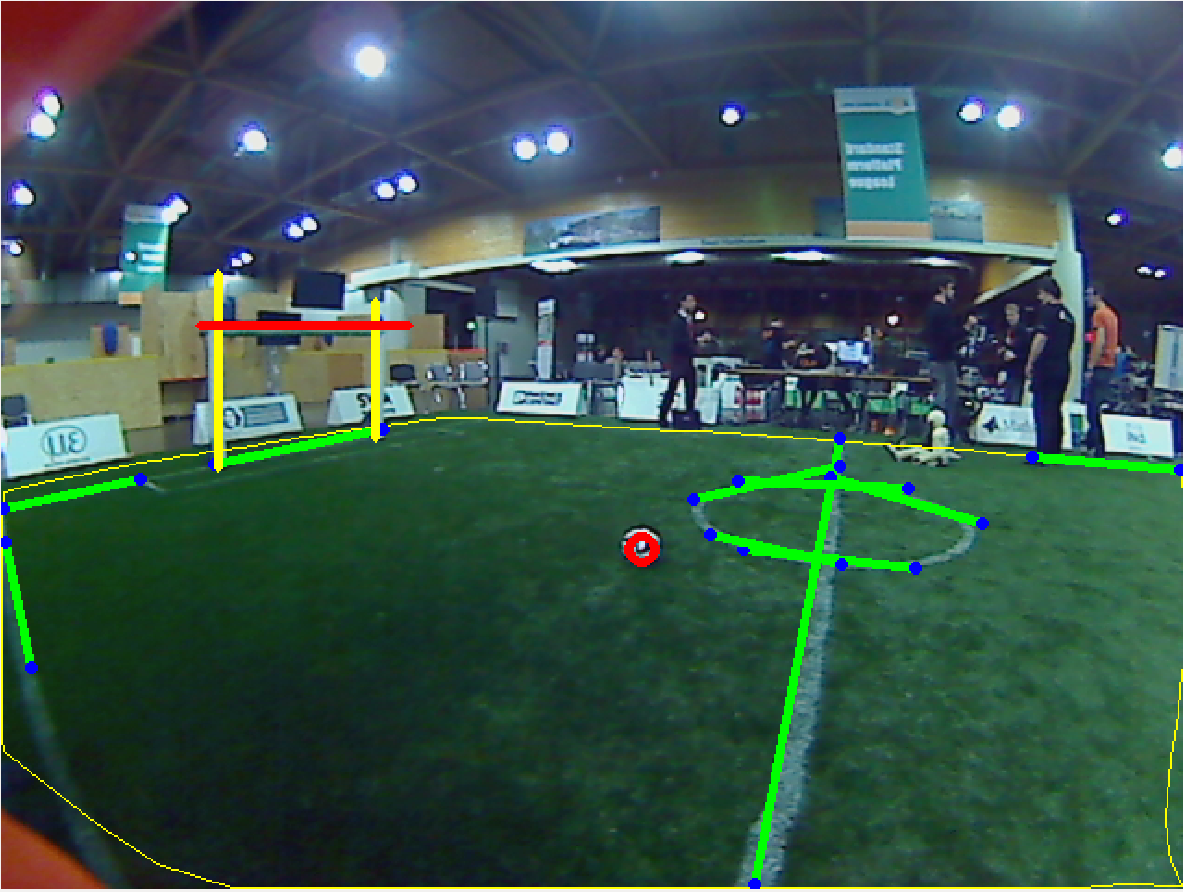}}
\caption{An image captured by the robot with ball (red circle), field line (green lines),
field boundary (thin yellow lines), goal post (thick yellow lines) and crossbar (red line)
detections annotated.}
\figlabel{vision_detections}
\vspace{-0.8em}
\end{figure}

\textbf{Fused Angle Deviation Pose Feedback:}
During regular stable open loop walking, the torso attitude undergoes 
regular limit cycles in both the lateral and sagittal directions. These trajectories 
are modelled in terms of fused pitch and fused roll using parameterised 
sinusoids that are functions of the gait phase. During walking, deviations to 
the expected trunk attitude are used to construct proportional, integral and 
derivative feedback terms. The proportional feedback term is obtained by 
applying deadband to the output of a mean filter that takes the fused angle 
deviations as its input, while the integral feedback term is obtained by 
applying an exponentially weighted integrator to the fused angle deviations. The 
derivative feedback term is obtained by applying a first order differentiating 
weighted line of best fit filter. This inherently provides enough smoothing on 
the differentiated output for feedback purposes. A matrix of gains is applied to 
the PID feedback vector, and the resulting feedback signals are applied to a set 
of stabilising mechanisms, namely phased offsets to the foot angle, leg angle, 
hip angle, arm angle and inverse kinematics centre of mass (CoM) position. This 
has the effect, for example, that the front of the support foot is pushed down 
into the ground, and the arms are moved backwards, when the robot is tipping 
forwards. The integral feedback terms work on a slower time scale, and 
essentially learn stabilising offsets to the central pose of the gait.

\textbf{Fused Angle Deviation Timing Feedback:}
While the fused angle pose feedback attempts to force the robot back into a 
regular gait limit cycle, the deviation timing feedback instead adapts the 
timing of the gait based on the deviation of the fused angle. It does this by 
adding proportional feedback from the current fused roll deviation to the rate 
of progression of the gait phase. Thus, if for example the robot starts to tip 
outwards over a support foot, the error in the fused roll causes the rate of the 
gait phase to slow down, effectively causing the robot to wait longer before 
attempting to land the next step. The opposite situation of an unbalance towards 
the swing foot would cause a speed up of the gait phase, causing the robot
to land its next step sooner. A region of deadband ensures that small 
natural deviations in the fused roll do not inject timing noise into the gait.

\textbf{Resulting Gait:}
With these feedback mechanisms in place, an omnidirectional 
gait was achieved on an artificial grass of \SI{32}{mm} blade length, with a 
walking speed of approximately \SI{21}{\centi\metre\per\second}. The cleats made 
a clear difference to the resulting stability of the gait, by reducing the 
sensitivity of the gait to the surface that the robot was walking on. On rare 
occasions, when the expected timing of the gait was too greatly disturbed by the 
grass during walking, the cleats got caught in the blades, causing further 
disturbances. In many cases however, even such situations were recoverable and 
did not necessarily lead to a fall.

\addtolength{\textheight}{-10mm}

\section{Vision System}
\seclabel{vision}

\begin{figure}[!t]
\parbox{\linewidth}{\centering\includegraphics[width=0.9\linewidth]{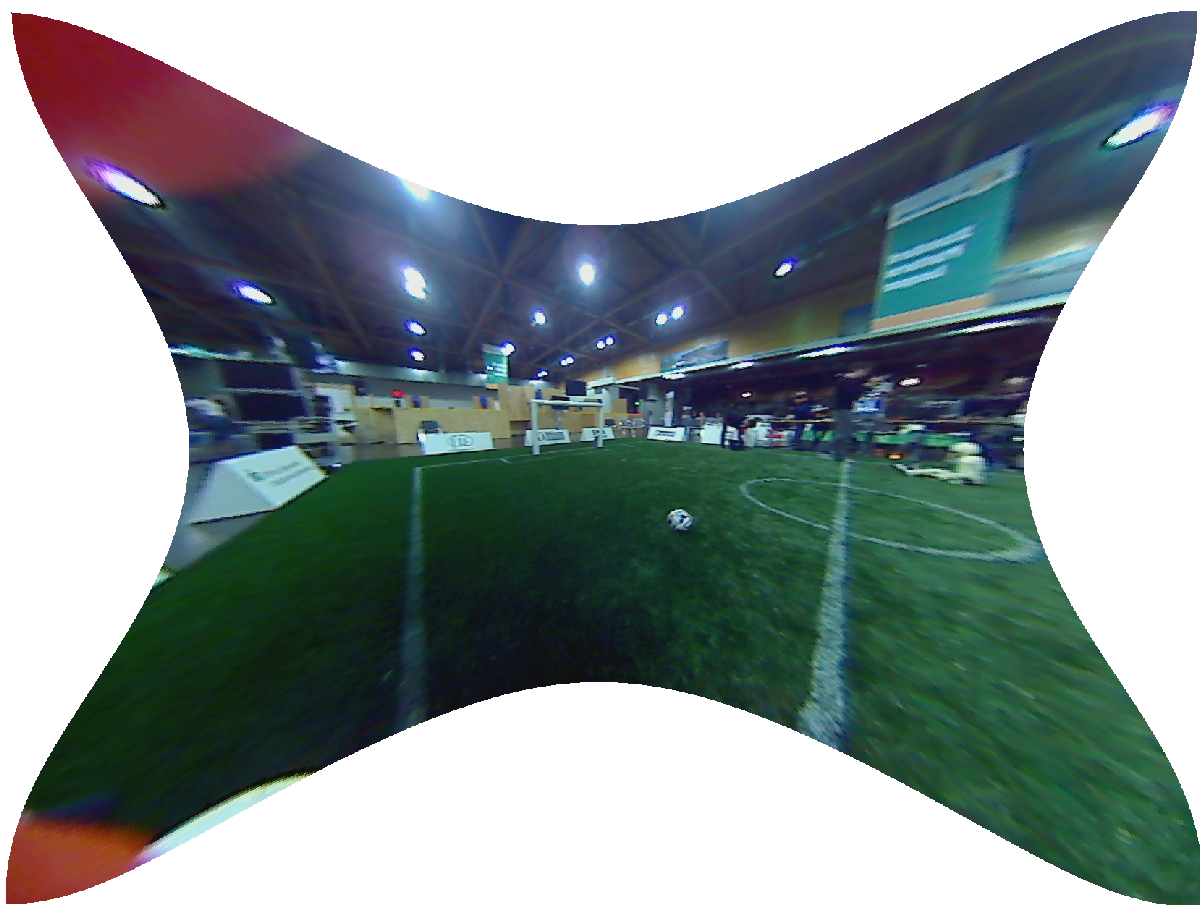}}
\caption{An image captured by the robot with the undistortion model applied.}
\figlabel{vision_distortion}
\end{figure}

The \iguhop is nominally fitted with a single 720p Logitech C905 camera behind 
its right eye. A second such camera can be mounted behind the left eye however, 
if needed for stereo vision. The camera is 
fitted with a wide-angle lens, yielding a field of view of 150\degree. The 
choice of lens was optimised from the first \nop prototype to increase the 
number of usable pixels and reduce the level of distortion, all without 
significantly sacrificing the effective field of view. The Video4Linux2 camera 
driver used in the ROS software nominally retrieves camera images at \SI{30}{Hz} 
in 24bpp BGR format at a resolution of $640\!\times\!480$. For further 
processing, including for example colour segmentation, the captured image is 
converted to the HSV colour space. In our target application of soccer, the 
vision processing tasks include field, ball, goal, line, centre circle and 
obstacle detection. A sample output of the detections in an image captured by 
the robot is shown in \figref{vision_detections}.

Due to the significant recent changes in the rules of the RoboCup Humanoid
League---relating largely to the visual features of the soccer field, including 
the introduction of a mostly white ball and white goals---new detection routines 
with reduced dependence on colour segmentation have been developed. For ball and 
goal detection, a histogram of oriented gradients (HOG) descriptor is applied in 
the form of a cascade classifier, with use of the AdaBoost technique. Sets of 
positive and negative samples are gathered directly from the robot camera, and 
used to train the classifier. Training with 20 stages, 400 positive and 700 
negative samples takes about 10 hours, and the resulting classifier has a 
success rate above 80\% on a walking robot, with very few false detections. As 
the HOG feature extraction is computationally expensive if applied to the whole 
image, some preprocessing steps are included to reject parts of the scene that 
do not contain objects of interest.

Due to the new artificial grass field in the humanoid league, line detection 
based on colour segmentation is no longer easily possible, mainly because the 
painted lines are no longer a clear white. Instead, an edge detector is used on 
the image brightness, followed by a probabilistic Hough Transform for line 
segment detection. The resulting line segments are filtered and merged based on 
their relative detected position in world coordinates. Shorter line segments are 
used to try to extract the field circle, and longer ones are used to update the 
localisation.

Due to the use of a wide-angle lens, a distortion model was required to be able 
to project points in the camera image into world coordinates. The distortion 
model used by OpenCV was adapted for the \iguhop, which uses two tangential and 
six radial distortion coefficients, in addition to the four standard intrinsic 
camera parameters. Due to incorrect performance of the OpenCV point undistortion 
function, a more efficient and accurate custom undistortion algorithm based on 
the Newton-Raphson method was designed and implemented. 
\figref{vision_distortion} illustrates the effect of undistortion on images 
captured by the robot.

\section{Conclusion}
\seclabel{conclusion}

The \iguhop represents a significant advancement over the previously released 
\nop robot towards a robust, affordable, versatile, and customisable open 
standard platform for humanoid robots in the child-sized range. The robot has 
proven to have many positive mechanical attributes, due largely to its modular 
design and advanced 3D printing manufacturing process that allows for near 
arbitrary flexibility in the design. The complete mechanical and electrical 
subsystems of the robot have been described in this paper, in addition to 
selected features of the developed ROS software, which remains a continual 
development effort. We have released the hardware and software open source to 
the community in the hope that it will benefit other research groups around the 
world, and that it will encourage them to also publish their results as a 
contribution to the open source community.

\section{Acknowledgements}

We would like to acknowledge the contributions of \igus GmbH to the project, in 
particular the management of Martin Raak towards the robot design and 
manufacture.

\bibliographystyle{IEEEtran}
\bibliography{IEEEabrv,ms}

\begin{thebibliography}{10}
\providecommand{\url}[1]{#1}
\csname url@samestyle\endcsname
\providecommand{\newblock}{\relax}
\providecommand{\bibinfo}[2]{#2}
\providecommand{\BIBentrySTDinterwordspacing}{\spaceskip=0pt\relax}
\providecommand{\BIBentryALTinterwordstretchfactor}{4}
\providecommand{\BIBentryALTinterwordspacing}{\spaceskip=\fontdimen2\font plus
\BIBentryALTinterwordstretchfactor\fontdimen3\font minus
  \fontdimen4\font\relax}
\providecommand{\BIBforeignlanguage}[2]{{%
\expandafter\ifx\csname l@#1\endcsname\relax
\typeout{** WARNING: IEEEtran.bst: No hyphenation pattern has been}%
\typeout{** loaded for the language `#1'. Using the pattern for}%
\typeout{** the default language instead.}%
\else
\language=\csname l@#1\endcsname
\fi
#2}}
\providecommand{\BIBdecl}{\relax}
\BIBdecl

\bibitem{Quigley2009}
M.~Quigley, K.~Conley, B.~P. Gerkey, J.~Faust, T.~Foote, J.~Leibs, R.~Wheeler,
  and A.~Ng, ``{ROS:} {A}n open-source robot operating system,'' in \emph{ICRA
  Workshop on Open Source Software}, 2009.

\bibitem{IguhopHardware}
\BIBentryALTinterwordspacing
{igus GmbH}. (2015, Oct) {igus Humanoid Open Platform Hardware CAD Data}.
  [Online]. Available: \url{https://github.com/igusGmbH/HumanoidOpenPlatform}
\BIBentrySTDinterwordspacing

\bibitem{IguhopSoftware}
\BIBentryALTinterwordspacing
{NimbRo}. (2015, Oct) {igus Humanoid Open Platform ROS Software}. [Online].
  Available: \url{https://github.com/AIS-Bonn/humanoid_op_ros}
\BIBentrySTDinterwordspacing

\bibitem{Gouaillier2009}
D.~Gouaillier, V.~Hugel, P.~Blazevic, C.~Kilner, J.~Monceaux, P.~Lafourcade,
  B.~Marnier, J.~Serre, and B.~Maisonnier, ``Mechatronic design of {NAO}
  humanoid,'' in \emph{Int. Conf. on Robotics and Automation}, 2009.

\bibitem{Ha2011}
I.~Ha, Y.~Tamura, H.~Asama, J.~Han, and D.~Hong, ``Development of open humanoid
  platform {DARwIn-OP},'' in \emph{SICE Annual Conf.}, 2011.

\bibitem{Lapeyre2014}
M.~Lapeyre, P.~Rouanet, J.~Grizou, S.~Nguyen, F.~Depraetre, A.~Le~Falher, and
  P.-Y. Oudeyer, ``{Poppy Project: Open-Source Fabrication of 3D Printed
  Humanoid Robot for Science, Education and Art},'' in \emph{{Digital
  Intelligence 2014}}, Sep 2014.

\bibitem{Hirai1998}
K.~Hirai, M.~Hirose, Y.~Haikawa, and T.~Takenaka, ``The development of {H}onda
  humanoid robot,'' in \emph{Int. Conf. on Rob. and Autom.}, 1998.

\bibitem{Kaneko2009}
K.~Kaneko, F.~Kanehiro, M.~Morisawa, K.~Miura, S.~Nakaoka, and S.~Kajita,
  ``Cybernetic human {HRP-4C},'' in \emph{Proceedings of 9th IEEE-RAS Int.
  Conf. on Humanoid Robotics (Humanoids)}, 2009, pp. 7--14.

\bibitem{Schwarz2012}
M.~Schwarz, M.~Schreiber, S.~Schueller, M.~Missura, and S.~Behnke, ``{NimbRo-OP
  Humanoid TeenSize Open Platform},'' in \emph{7th Workshop on Humanoid Soccer
  Robots, Int. Conf. on Humanoid Robots}, 2012.

\bibitem{Allgeuer2013a}
P.~Allgeuer, M.~Schwarz, J.~Pastrana, S.~Schueller, M.~Missura, and S.~Behnke,
  ``A {ROS}-based software framework for the {NimbRo-OP} humanoid open
  platform,'' in \emph{8th Workshop on Humanoid Soccer Robots, Int. Conference
  on Humanoid Robots}, 2013.

\bibitem{Schwarz2013}
M.~Schwarz, J.~Pastrana, P.~Allgeuer, M.~Schreiber, S.~Schueller, M.~Missura,
  and S.~Behnke, ``{Humanoid TeenSize Open Platform NimbRo-OP},'' in
  \emph{Proceedings of 17th RoboCup International Symposium}, Eindhoven,
  Netherlands, 2013.

\bibitem{Allgeuer2013}
P.~Allgeuer and S.~Behnke, ``Hierarchical and state-based architectures for
  robot behavior planning and control,'' in \emph{Proceedings of 8th Workshop
  on Humanoid Soccer Robots, IEEE-RAS Int. Conference on Humanoid Robots},
  Atlanta, USA, 2013.

\bibitem{Schwarz2013a}
M.~Schwarz and S.~Behnke, ``Compliant robot behavior using servo actuator
  models identified by iterative learning control,'' in \emph{Proceedings of
  17th RoboCup Int. Symposium}, Eindhoven, Netherlands, 2013.

\bibitem{Felis2013}
\BIBentryALTinterwordspacing
M.~Felis. (2015, Jun) {Rigid Body Dynamics Library}. [Online]. Available:
  \url{http://rbdl.bitbucket.org/}
\BIBentrySTDinterwordspacing

\bibitem{Allgeuer2015}
P.~Allgeuer and S.~Behnke, ``{F}used {A}ngles: {A} representation of body
  orientation for balance,'' in \emph{Int. Conf. on Intelligent Robots and
  Systems (IROS)}, Hamburg, Germany, 2015.

\bibitem{Allgeuer2014}
P.~Allgeuer and S.~Behnke, ``Robust sensor fusion for biped robot attitude
  estimation,'' in \emph{Proceedings of 14th IEEE-RAS Int. Conference on
  Humanoid Robotics (Humanoids)}, Madrid, Spain, 2014.

\bibitem{Behnke2006}
S.~Behnke, ``Online trajectory generation for omnidirectional biped walking,''
  in \emph{Proceedings of 2006 IEEE International Conference on Robotics and
  Automation}, Orlando, USA, 2006.

\bibitem{Missura2013a}
M.~Missura and S.~Behnke, ``Self-stable omnidirectional walking with compliant
  joints,'' in \emph{Proceedings of 8th Workshop on Humanoid Soccer Robots,
  Int. Conf. on Humanoid Robots}, 2013.

\end{thebibliography}

\end{document}